\documentclass[10pt,twocolumn,letterpaper]{article}

\usepackage{wacv_arxiv}
\usepackage{times}
\usepackage{epsfig}
\usepackage{graphicx}
\usepackage{amsmath}
\usepackage{amssymb}

\usepackage{float}
\usepackage{caption}
\usepackage{subcaption}

\usepackage{verbatim}
\usepackage{paralist}
\usepackage{nccmath}
\DeclareMathOperator*{\argmax}{argmax} 

\def\wacvPaperID{-} 

\wacvfinalcopy 

\ifwacvfinal
\def\assignedStartPage{1} 
\fi

\ifwacvfinal
\usepackage[breaklinks=true,bookmarks=false]{hyperref}
\else
\usepackage[pagebackref=true,breaklinks=true,colorlinks,bookmarks=false]{hyperref}
\fi

\ifwacvfinal
\else
\fi

\usepackage{caption} 
\begin{document}

\title{Evaluating the Robustness of Semantic Segmentation for Autonomous Driving against Real-World Adversarial Patch Attacks\thanks{This work has been submitted to the IEEE for possible publication. Copyright may be transferred without notice, after which this version may no longer be accessible.}}

\author{Federico Nesti\thanks{Equal contribution.}~, ~Giulio Rossolini\footnotemark[2]~, ~Saasha Nair,  ~Alessandro Biondi, ~ and Giorgio Buttazzo\\
Department of Excellence in Robotics \& AI, Scuola Superiore Sant'Anna\\
{\tt\small <name>.<surname>@santannapisa.it}
}

\maketitle

\begin{abstract}
Deep learning and convolutional neural networks allow achieving impressive performance in computer vision tasks, such as object detection and semantic segmentation (SS). 
However, recent studies have shown evident weaknesses of such models against adversarial perturbations. In a real-world scenario instead, like autonomous driving, more attention should be devoted to \textit{real-world adversarial examples} (RWAEs), which are physical objects (e.g., billboards and printable patches) optimized to be adversarial to the entire perception pipeline.
This paper presents an in-depth evaluation of the robustness of popular SS models by testing the effects of both digital and real-world adversarial patches. These patches are crafted with powerful attacks enriched with a novel loss function. Firstly, an investigation on the Cityscapes dataset is conducted by extending the Expectation Over Transformation (EOT) paradigm to cope with SS.
Then, a novel attack optimization, called scene-specific attack, is proposed. Such an attack leverages the CARLA driving simulator to improve the transferability of the proposed EOT-based attack to a real 3D environment. 
Finally, a printed physical billboard containing an adversarial patch was tested in an outdoor driving scenario to assess the feasibility of the studied attacks in the real world.
Exhaustive experiments revealed that the proposed attack formulations outperform previous work to craft both digital and real-world adversarial patches for SS. At the same time, the experimental results showed 
how these attacks are notably less effective in the real world, hence 
questioning the practical relevance of adversarial attacks to SS models for autonomous/assisted driving.
   
\end{abstract}

\vspace{-1.5em}
\section{Introduction}
\vspace{-0.5em}
The rise of deep learning unlocked unprecedented performance in several scientific areas \cite{dl_appl}. 
Convolutional neural networks \cite{krizhevsky_imagenet_2017} (CNNs) yielded super-human performance for many different computer vision tasks, such as image recognition \cite{2015arXiv151207108G}, object detection \cite{2015arXiv150601497R} \cite{2015arXiv150602640R}, 
 and image segmentation \cite{minaee_image_2020}. Image segmentation, and semantic segmentation (SS) in particular, is used in autonomous driving perception pipelines \cite{siam2018comparative}, mainly for object detection \cite{minaee_image_2020}. 


Despite their high performance, CNNs are prone to adversarial attacks \cite{silva_opportunities_2020}. Most of the literature on adversarial attacks focuses on directly manipulating the pixels of the whole image, hence making the assumption that the attacker has control over the digital representation of the environment obtained by the on-board cameras. This kind of unsafe inputs are called \emph{digital} adversarial examples.

 Although such digital attacks do not transfer well into the real world, they continue to be used to evaluate the robustness of models in safety-critical systems \cite{survey_trust, 
bar_vulnerability_2021, 
metzen_universal_2017}.
\emph{Real-world} adversarial examples (RWAEs), on the other hand, are physical objects that can be placed in the field of view of a camera, such that the resulting image acts as an adversarial example for the neural network under attack \cite{kurakin_adversarial_2017}. Thus, RWAEs can induce errors in neural networks without requiring the attacker to access the digital representation of the image, thereby making them a more realistic and dangerous threat to safety-critical systems.



\begin{figure}[!t]
     \centering
     \begin{subfigure}{0.155\textwidth}
         \centering
         \includegraphics[width=\textwidth]{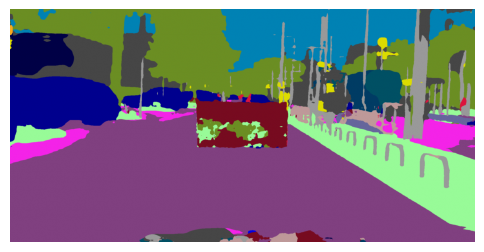}
         \vspace{-1.5em}
         \caption{}
         \label{fig:carla_exps_a}
     \end{subfigure}
     \begin{subfigure}{0.155\textwidth}
         \centering
         \includegraphics[width=\textwidth]{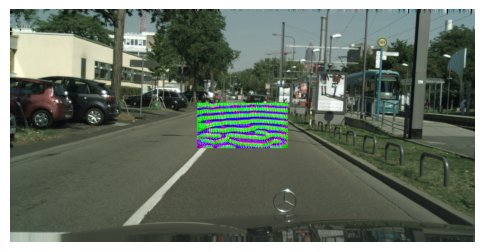}
         \vspace{-1.5em}
         \caption{}
         \label{fig:carla_exps_b}
     \end{subfigure}
     \begin{subfigure}{0.155\textwidth}
         \centering
         \includegraphics[width=\textwidth]{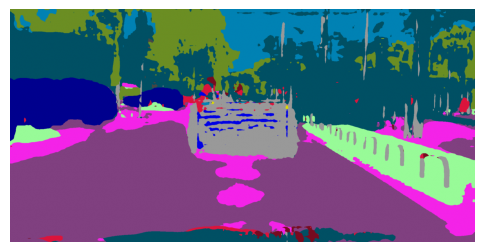}
         \vspace{-1.5em}
         \caption{}
         \label{fig:carla_exps_c}
     \end{subfigure}
     \begin{subfigure}{0.155\textwidth}
         \centering
         \includegraphics[width=\textwidth]{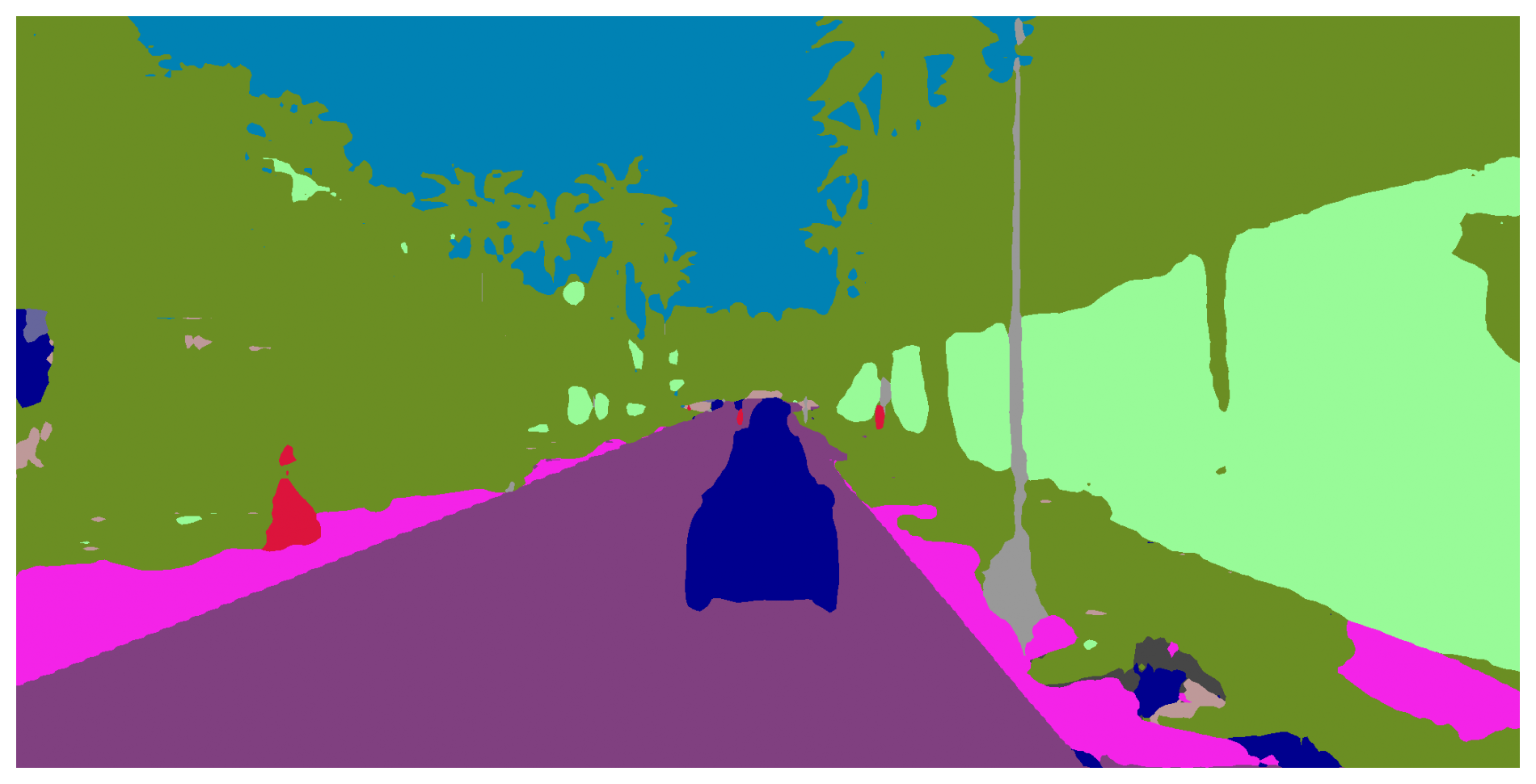}
         \vspace{-1.5em}
         \caption{}
         \label{fig:carla_exps_d}
     \end{subfigure}
     \begin{subfigure}{0.155\textwidth}
         \centering
         \includegraphics[width=\textwidth]{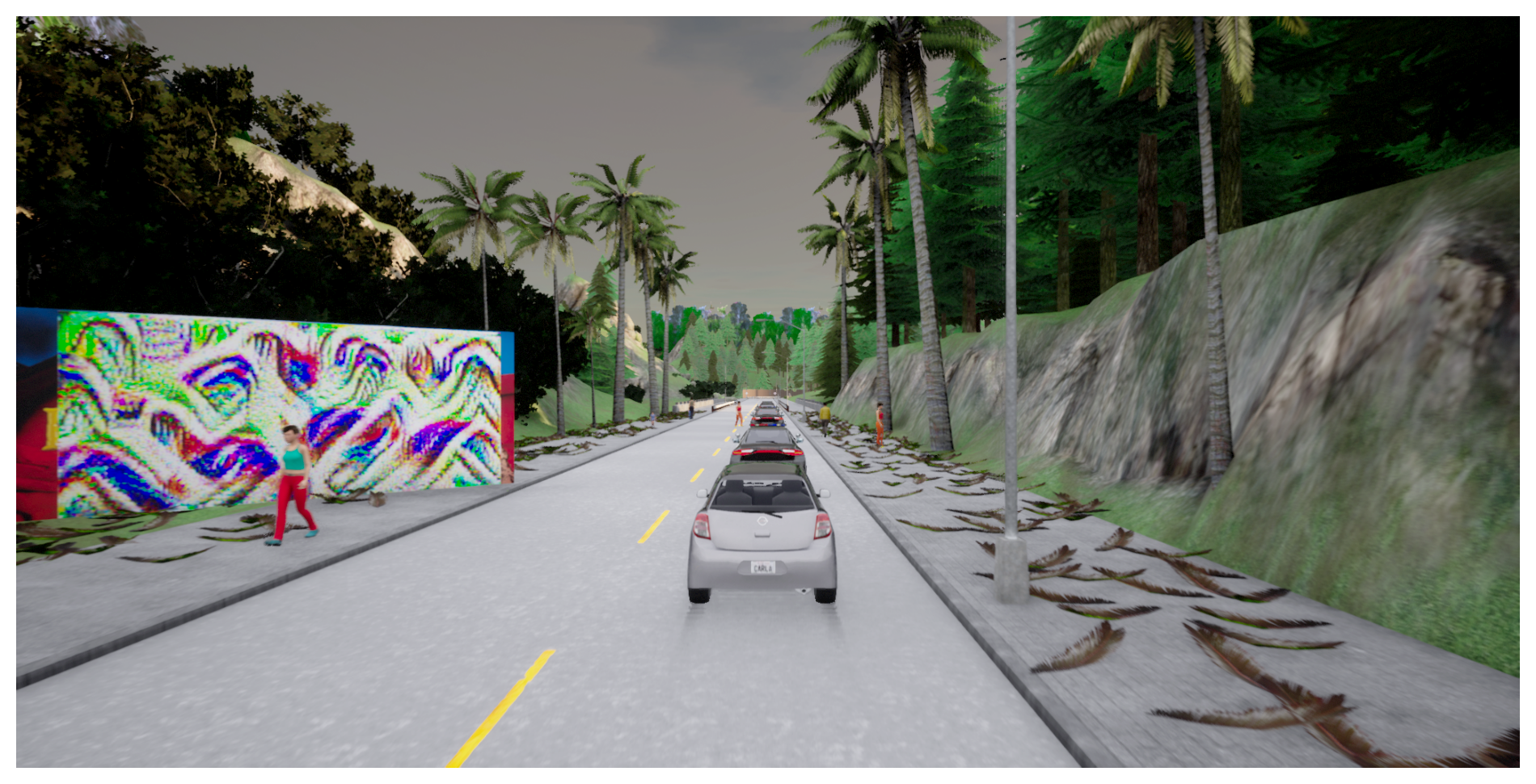}
         \vspace{-1.5em}
         \caption{}
     \end{subfigure}
     \begin{subfigure}{0.155\textwidth}
         \centering
             \includegraphics[width=\textwidth]{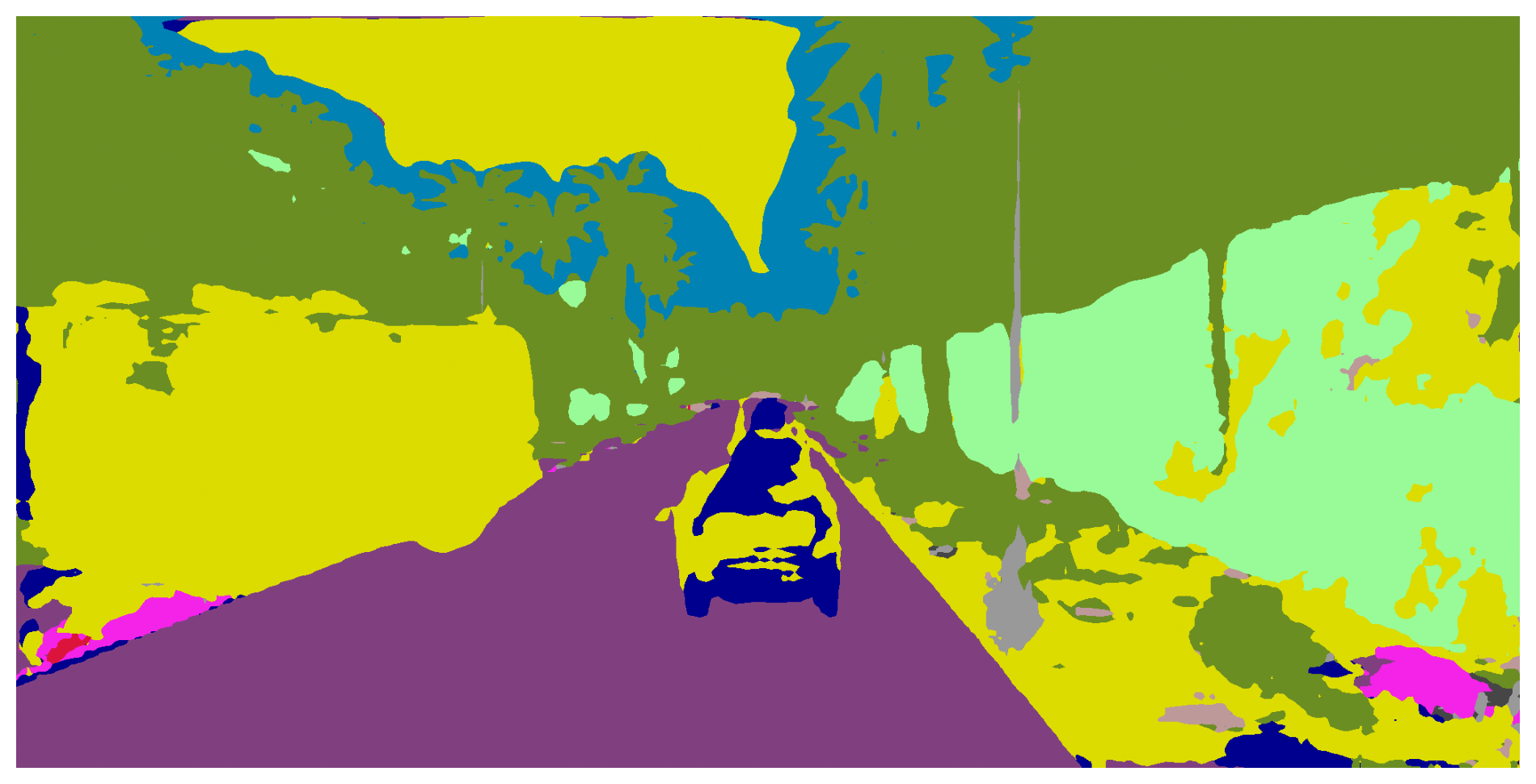}
             \vspace{-1.5em}
         \caption{}
     \end{subfigure}
        \vspace{-1em}
        \caption{Proposed adversarial patches on Cityscapes \cite{DBLP:conf/cvpr/CordtsORREBFRS16} (b) and CARLA Simulator \cite{2017arXiv171103938D} (e); (c/f) show the corresponding SS predicted by BiSeNet \cite{bisenet_paper}; (a/d) show the corresponding predictions obtained using random patches instead of adversarial ones.}
        \vspace{-1em}
        \label{fig:intro_exps}
\end{figure}

\smallskip
\noindent
\textbf{This paper.}
This work focuses on RWAEs, as they represent a potential threat to tasks in autonomous driving today. Although the effects of RWAEs have been studied extensively in the literature for classification and object detection, those on SS remain relatively unexplored. However, SS is an integral part of autonomous driving pipelines~\cite{siam2018comparative}. Thus, this paper examines various state-of-the-art models for real-time SS aiming at benchmarking their robustness to RWAEs in autonomous driving scenarios.

Of the several types of RWAEs proposed in the literature \cite{sun_survey_2018}, the form of attack used in this paper is adversarial patches \cite{brown_adversarial_2018}. This is because attacks that perturb the whole image are not practically feasible in the real world. Conversely, such patches can be easily printed and attached to any visible 2D surface in the driving environment, such as billboards and road signs, thus making them a simple, yet effective attack strategy.

The paper starts by recognizing the shortcomings of the standard cross-entropy loss for optimizing adversarial patches for SS. Thus, an extension to the cross-entropy loss is proposed and integrated in all the performed attacks. This extension forces the optimization to focus on pixels that are not yet misclassified, thus obtaining patches that are more powerful compared to those generated with the standard cross-entropy-based setting~\cite{nakka_indirect_2019}.

Following this rationale, the robustness of real-time SS models to RWAEs attacks is benchmarked. 
The paper starts by first examining the \textbf{case of driving images}, crafting adversarial patches on the Cityscapes dataset \cite{DBLP:conf/cvpr/CordtsORREBFRS16}, a popular benchmark of high-resolution images of urban driving.
Robust real-world patches are crafted by following the Expectation Over Transformation (EOT) \cite{pmlr-v80-athalye18b} paradigm, which has been extended in this work to attack SS models. 
Furthermore, a comparison against non-robust patches (without EOT) is presented to question their effectiveness on driving scenes. 

Another set of experiments targeted a \textbf{virtual 3D scenario}, for which a stronger adversarial attack is presented and tested. The proposed \emph{scene-specific attack}, defined in Section \ref{subsection:scene-specific-approach}, is a more practical tool for crafting adversarial patches in a realistic autonomous driving scenario. It assumes that the attacker is interested in 
targeting an autonomous driving scene at a particular corner of a specific town, where information about the position of the attackable 2D surface (in our case, a billboard) is available. To satisfy such requirements we developed and tested this attack using the CARLA simulator, which provides all the needed geometric information. These experiments include a comparison with the EOT-based and non-robust patches, performed by importing them into the CARLA world and placing them on billboards to simulate a realistic study. 

Figure \ref{fig:intro_exps} provides some examples of the effect of our patches on Cityscapes and CARLA.

The last set of experiments were conducted on a \textbf{real-world driving scenario}, which required collecting a dataset within the city, optimizing a patch on it, physically printing said patch on a billboard, and finally evaluating SS models on images containing the printed patch.

To the best of our knowledge, this work represents the first exhaustive evaluation of the robustness of SS models against RWAEs for autonomous driving systems. The results of the experiments state important observations that should be taken into consideration while evaluating the trustworthiness of SS models in autonomous driving. First, they demonstrate that non-robust patches are not good candidates for assessing the practical robustness of an SS model to adversarial examples. Indeed, while they proved to be effective in attacking images related to driving scenes (from Cityscapes), they do not induce any real-world adversarial effect when crafted and tested in a virtual 3D world (based on CARLA). 
Conversely, robust patches, crafted with EOT or the proposed scene-specific approach, resulted to be less effective than non-robust ones on Cityscapes images, but were capable to accomplish the attack in both virtual 3D world and the real world.
Nevertheless, their effectiveness in the latter two cases still resulted to be quite limited, hence questioning the practical relevance of RWAEs.


In summary, the paper makes the following contributions:
\begin{compactitem}
 \item It proposes an extension to the pixel-wise cross-entropy loss to enable crafting strong patches for the semantic segmentation setting.
  \item It proposes a novel technique for crafting adversarial patches for autonomous driving scenarios that utilize geometric information of the 3D world.
 \item It finally reports an extensive evaluation of RWAE-based attacks on a set of real-time semantic segmentation models using data from the Cityscapes dataset, CARLA, and the real world.
\end{compactitem}

The remainder of this paper is organized as follows: Section \ref{s:related} provides a brief overview of related work existing in the literature, Section \ref{s:method} formalizes the proposed loss function, pipeline, and attack strategy, Section \ref{s:exp} reports the experimental results, and Section \ref{s:conclusions} states the conclusions and proposes ideas for future work.

\section{Related Work} \label{s:related}


Szegedy et al. \cite{DBLP:journals/corr/SzegedyZSBEGF13} showed that small well-crafted perturbations when added to the input image were sufficient to fool strong classification networks. 
\cite{bar_vulnerability_2021, 
metzen_universal_2017, 
DBLP:conf/iccv/XieWZZXY17,  
DBLP:journals/access/KangSDG20, 
2019arXiv190805005K,  
arnab_robustness_nodate, 
nakka_indirect_2019,
shen2019advspade} have studied such attacks for the specific use case of fooling SS models. However, these attacks directly manipulate the pixels of the image. Although such digital perturbations provide a convenient way to provide benchmarks in research, they do no extend well to real-world applications.


A more realistic threat model led to the introduction of RWAEs by Kurakin et. al. \cite{kurakin_adversarial_2017}. The attacker here is assumed to be able to craft adversarial pictures in the physical world, without the ability to manipulate the digital representation of inputs to the neural network. However, this work did not account for factors that affect images of objects in the real-world (e.g., varying viewpoints from which input images could be captured, changes in lighting conditions and so on). Athalye et. al. \cite{pmlr-v80-athalye18b} address this issue by introducing the EOT algorithm. EOT accounts for such factors in the optimisation by modeling them as a distribution of transformation functions applied to the adversarial input. These transformations can be in the form of rotation, scaling, noise, brightening and so on. Then, the idea is to optimize the loss function in expectation across the range of selected transformation functions.

The EOT formulation led to the development of \emph{adversarial patches}, introduced by Brown et al.~\cite{brown_adversarial_2018} to fool image classifiers. They are robust, localized, image-agnostic perturbations, crafted with the EOT paradigm, capable of fooling neural networks when placed within the input scene or added digitally on images. 

Although extensive prior work exists to construct such physical attacks for classification~\cite{brown_adversarial_2018, sharif_accessorize_2016, eykholt_robust_2018}, object detection \cite{DBLP:conf/eccv/WuLDG20, wu_physical_2020, lee_physical_2019, zhang_camou_2019}, optical flow \cite{2019arXiv191010053R}, LiDAR object detection \cite{Tu2020PhysicallyRA}, and depth estimation \cite{2020arXiv201003072Y}, only a few focus on autonomous driving tasks, since testing the adversarial robustness is more challenging, as it requires controlling the 3D outdoor environments. 
Other works \cite{zhang_camou_2019, wu_physical_2020} have shown CARLA to be a viable solution in alleviating this issue by crafting and evaluating adversarial situations in virtual 3D environments. This paper also heavily relies on CARLA to evaluate how the optimized adversarial patches translate to a 3D world.


The work closest to ours is the one by Nakka et. al. \cite{nakka_indirect_2019}, who attempted to fool a variety of SS models via local attacks (i.e., creating pixel perturbation in a specific area of the image). Despite the attacks being local, the objective of their study was not to evaluate the robustness to real-world attacks, which is instead the main focus of this paper.
To the best of our knowledge, such a study is missing in the literature for the case of SS models, which represent essential components in an autonomous driving perception pipeline \cite{siam2018comparative}. 



Additionally, this paper also improves the loss function used for generating patches. 
Section \ref{subsection:loss} presents a more general formulation of the cross-entropy loss for the SS setting, designed to optimize more powerful and effective adversarial examples, while all the previously mentioned papers use the standard pixel-wise cross-entropy loss. 



\section{Attack Formulation} \label{s:method}

This section presents the design of adversarial patches for semantic segmentation (SS), starting with a short recap of the basic notions behind SS. The patch optimization scheme for both the EOT-based and the scene-specific attacks is then presented. Finally, the proposed loss function is introduced.


\subsection{Background on SS}
An image with height $H$ and width $W$ can be represented as $x \in [0,1]^{H \times W \times C}$, 
where $C$ is the number of channels.
An SS model returns $f(x) \in [0,1]^{H \times W  \times N_c}$, where $N_c$ is the number of classes.
This output represents the predicted class-probability scores associated to each image pixel.
In particular, $f_{i}^{j}(x) \in [0,1]$ denotes the predicted probability score for the $i$-th pixel of the image corresponding to the class $j$.
Consequently, the predicted semantic segmentation $SS(x) \in \mathbb{N}^{H \times W} $ is computed by extracting those classes with the highest probability score in each pixel: $SS(x) = { \argmax_{j \in \{1,...,N_c\} }}{f_i^j(x) } ~, \forall i \in \{1,...,H \times W\}$.

The ground truth for the SS of $x$ is defined as $y \in \mathbb{N}^{H \times W}$, and assigns the correct class (in $\{1,...,N_c\}$) to each pixel.
The performance of the SS models is evaluated by computing the cross-entropy loss $\mathcal{L}_{CE}(f_i(x), y_i) = -log({f_i^{y_i}(x)})$. Thus, for each pixel $i$, the model's prediction $f_i$ is compared against the ground truth class $y_i$.

\subsection{Patch-based attack pipeline}
\begin{figure}[!t]
\centering
\makebox[\columnwidth]{\includegraphics[scale=0.31]{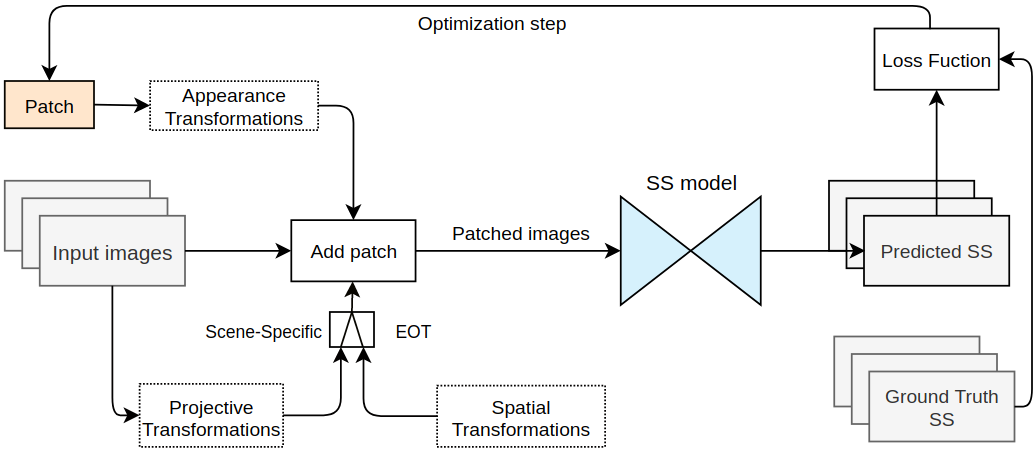}}
\caption{Outline of the proposed approach for crafting both the EOT-based and the scene-specific patches.}
\vspace{-1em}
\label{fig:attack_scheme}
\end{figure}

Both the EOT-based and the scene-specific attacks share a similar pipeline, which is explained in the following paragraph and illustrated in Figure~\ref{fig:attack_scheme}. 

An adversarial patch of height $\tilde{H}$ and width $\tilde{W}$ is denoted as $\delta \in [0,1]^{\tilde{H} \times \tilde{W} \times C}$, where 
$\tilde{H}<H$ and $\tilde{W}<W$. This patch is then added to the original image $x$ to obtain a patched image $\tilde{x}$.
Thus, the output of the SS model on this patched image would now be $f(\tilde{x})$.

The attacks considered are both untargeted. This means that the objective is to maximize a certain loss function $\mathcal{L}$, without forcing the classification of pixels towards any specific class.



Inspired by the EOT method \cite{pmlr-v80-athalye18b}, the idea is to find an optimal patch $\delta^*$ (starting from a random patch) that maximizes the loss $\mathcal{L}$ for all the patched images in expectation according to the distribution of transformations used to apply the patch $\delta$ on the image set $\mathbf{X}$.

Formally, we need to define:
\begin{compactitem}
    \item A \textbf{set of appearance-changing transformations $\Gamma_a$}, for instance changes in illumination (brightness, contrast) and noise (uniform or gaussian). These transformations are directly applied to the patch, so obtaining a transformed patch $\zeta_a(\delta)$, where $\zeta_a\in\Gamma_a$. They are used to make the patch robust to illumination changes and acquisition noise. 
    \item A \textbf{patch placement function $\eta$} that defines which portion of the original image $x$ is occupied by the patch. This is the only part of the pipeline that differs between the two proposed attacks, and is discussed further in the following subsections.
    \item A \textbf{patch application function $g(x,  \zeta_a(\delta), \eta)$} that replaces a certain area of $x$ with $\zeta_a(\delta)$ according to $\eta$ and returns the patched image $\tilde{x}$.
\end{compactitem}

These functions are sufficient to define both the \textit{EOT-based} attack, which uses randomized spatial transformations to place the patch onto the image, and the \textit{scene-specific} attack, which uses a precise projective transformation to enhance the accuracy of the patch placement.

\subsection{EOT-based patch attack}
The classic EOT-based attack, as in previous work \cite{brown_adversarial_2018}~\cite{2019arXiv191011099X}, 
uses a set of combinations of spatial transformations $\Gamma$, including translation and scaling, from which the patch placement function $\eta$ is selected.

The parameters for each transformation are randomized within a pre-defined range. Section~\ref{s:exp} provides a more detailed explanation of the set of transformations used.
The optimal patch is then defined as
\begin{equation}
    \delta^* = \argmax_{\delta} ~ \mathbb{E}_{x \in \mathbf{X}, \zeta_a \in \mathbf{\Gamma_a}, \eta \in \mathbf{\Gamma}} ~ 
    \mathcal{L}(f(\tilde{x}), y) ~
\end{equation}



In practice, the optimal patch is computed via an iterative optimization process.
At each iteration $t$, the pixels values of the patch are modified in the direction of the gradient of the loss function computed with respect to the patch:
\useshortskip
\begin{equation}
    \delta_{t+1} = \mathrm{clip}_{[0,1]} \left(\delta_{t} + \epsilon \cdot\sum_{x \in \mathbf{X}} \sum_{\substack{\zeta_a\in\mathbf{\Gamma_a}\\ \eta\in \mathbf{\Gamma}}}\nabla_{\delta_{t}} \mathcal{L}(f(\tilde{x}), y) \right ),
\end{equation} \label{e:opt}
\noindent where $\epsilon$ represents the step size.
$\mathcal{L}$ consists of a weighted sum of multiple loss functions. The adversarial patch effect is obtained through the optimization of the adversarial loss $\mathcal{L}_{adv}$ (discussed further in subsection \ref{subsection:loss}). Additionally, to ensure that the patch transfers well to the real world, two losses are added to account for the \emph{physical realizability} of the patch
(see supplementary material\footnote{The paper includes references to additional material, that can be provided upon request.}): smoothness loss $\mathcal{L}_S$ and non-printability score $\mathcal{L}_N$.

\subsection{Scene-specific patch attack} \label{subsection:scene-specific-approach}
To provide a more realistic approach for autonomous driving environments, this work proposes an alternative attack methodology that exploits the geometrical information provided by the CARLA Simulator~\cite{2017arXiv171103938D}. 

Here, the key assumption is the availability of an attackable 2D surface, e.g., a billboard, with a fixed location in close proximity to the road.
The CARLA simulator features the possibility to extract camera extrinsic and intrinsic matrices (details in supplementary material), and the pose of the attackable surface. The billboard-to-image transformations can be computed using a 3D roto-translation composition, which allows the patch to be warped accordingly, thus obtaining higher precision in applying the patch to the attack surface.


This attack uses the same optimization pipeline as before, with one major difference: instead of placing the patch randomly, as in the previous attack, correct projective transformations are used to determine the placement of the patch on the attackable surface. Hence, $\eta$ is no longer randomized, but is computed for each image in the dataset.

This method allows crafting precise attacks that are optimised for the region of the town that the attacker is interested in. The attacker would need to collect several images, from different viewpoints, of the desired attackable surface, along with the corresponding intrinsic and extrinsic matrices. This approach to image collection also implies that EOT is no longer needed for patch placement, thereby simplifying the optimization process.

The downside of this approach is that a digital representation of the target scene is required to accurately capture the required matrices. Although CARLA provides the possibility to import cities via OpenStreetMaps (\url{https://www.openstreetmap.org/}), it requires some amount of manual effort to properly model 3D meshes to include objects in this simulated world. These objects need to be properly designed to ensure that the patches optimised in simulation transfer well to the real-world.
%
Although this paper does not investigate CARLA-to-real-world transfer issues, future work will address this problem to improve the proposed methodology and adapt it for real-world attacks.
Section \ref{s:exp} provides a comparison of this method against the EOT-based attack.


\subsection{Proposed loss function}\label{subsection:loss}

Cross-entropy (CE) is a popular choice as adversarial loss. Pixel-wise CE has been shown to work well when crafting an untargeted digital attack (i.e., by directly adding a perturbation $r$ to the pixels of a digital image)~\cite{nakka_indirect_2019}~\cite{bar_vulnerability_2021}.
This is formulated as: $\mathcal{L}_{adv}(f(x+r), y) = \frac{1}{|\mathcal{N}|}\cdot \sum_{i \in \mathcal{N}} \mathcal{L}_{CE}(f_i(x+r), y_i)$, where $\mathcal{N} = \{1,...,H \times W\}$ denotes the whole set of pixels in $\tilde{x}$.
However, modifications can be introduced to this formulation to allow crafting stronger attacks for fooling SS models.


Following previous notation, let $\mathcal{\tilde{N}} =\{1,..., \tilde{H} \times \tilde{W} \}  \subseteq \mathcal{N}$ denote only the pixels that correspond to the patch $\delta$. Then, $\mathbf{\Upsilon}$ defines a subset of image pixels that do not belong to the patch and are still predicted correctly by the model with respect to the trusted ground truth label $y$:
\useshortskip
\begin{equation}
\mathbf{\Upsilon} = \{ i \in  \mathcal{N} \setminus \mathcal{\tilde{N}} \quad | \quad SS_i(\tilde{x}) = y_i \}.
\end{equation}

Using $\mathbf{\Upsilon}$, the previous pixel-wise CE loss computed on $\mathcal{N} \setminus \mathcal{\tilde{N}}$ can be split into two distinct terms:
\useshortskip
\begin{equation}
    \mathcal{L}_M^{\tilde{x}}  =  \sum_{i \in \mathbf{\Upsilon}} \mathcal{L}_{CE}(f_i(\tilde{x}), y_i), \quad
    \mathcal{L}_{\overline{M}}^{\tilde{x}}  =  \sum_{i \notin \mathbf{\Upsilon}} \mathcal{L}_{CE}(f_i(\tilde{x}), y_i) ~.
\end{equation}

$\mathcal{L}^{\tilde{x}}_M$ describes the cumulative CE for those pixels that have been misclassified with respect to the ground truth $y$, while $\mathcal{L}^{\tilde{x}}_{\overline{M}}$ refers to all the others. 

Note that both $\mathcal{L}^{\tilde{x}}_M$ and $\mathcal{L}^{\tilde{x}}_{\overline{M}}$ do not consider pixels of the patch, which have been discarded to focus the optimization on attacking portions of the image away from the patch.
By computing these separate contribution to the total loss, we avoid that the contribution of the non-misclassified pixels gets obscured by the other term, which is a problem we found during preliminary tests.
Hence, the adversarial loss function gradient is redefined as follows:
\useshortskip
\begin{equation}
\nabla_{\delta} \mathcal{L}(f(\tilde{x}), y) = \gamma \cdot \frac{ \nabla_{\delta} \mathcal{L}_M^{\tilde{x}}}{||\nabla_{\delta} \mathcal{L}_M^{\tilde{x}}||_{2}} + (1-\gamma) \cdot \frac{ \nabla_{\delta} \mathcal{L}_{\overline{M}}^{\tilde{x}}}{||\nabla_{\delta} \mathcal{L}_{\overline{M}}^{\tilde{x}})||_{2}}  ~,
\end{equation}

\noindent where $\gamma \in [0,1]$ is a parameter that determines whether the optimization should focus on decreasing the number of correctly classified pixels or improving the adversarial strength for the currently misclassified pixels.
The rationale of $\gamma$ is to provide an empirical balancing between the importance of $\mathcal{L}_M$ and $\mathcal{L}_{\overline{M}}$ at each iteration $t$ depending on the number of pixels not yet misclassified.

Moreover, an adaptive value of $\gamma = \frac{|\Upsilon|}{|\mathcal{N} \setminus \mathcal{\tilde{N}}|}$ has been proposed to provide an automatic tuning of $\gamma$ at each iteration. 
The idea is to initially focus on boosting the number of misclassified points. 
Over time, as this number increases, the focus of the loss function gradually shifts toward improving the adversarial strength of the patch on these wrongly classified pixels.


Section \ref{s:exp} provides an extensive analysis of the proposed loss function by comparing multiple values of $\gamma$ with the standard pixel wise CE measured both on $\mathcal{N} \setminus \mathcal{\tilde{N}}$ and $\mathcal{N}$ (which is used by \cite{nakka_indirect_2019}), suggesting that our formulation is indeed more general and effective for this kind of attack.

\section{Experimental results} \label{s:exp}
This section describes the experimental setup and the results achieved with the proposed attacks. 
%
First, the proposed loss function is evaluated for different values of $\gamma$ comparing its effectiveness against the standard pixel-wise CE, showing that it is a better alternative for this kind of problems. Following this, the results of patches crafted with and without EOT are presented on the Cityscapes dataset. 

%
Subsequently, the results obtained with the scene-specific attack on three CARLA-generated datasets are presented. These results are compared against the EOT-based attack to show the improved effectiveness of this formulation.
Finally, some preliminary results of real-world adversarial patches are presented.
A more detailed analysis of all models tested against these attacks can be found in the supplementary material.


\subsection{Experimental setup} \label{s:exp_setup}
The experiments were performed on a set of 8 NVIDIA-A100 GPUs, while the CARLA simulator was run on a system powered by an Intel Core i7 with 12GB RAM and a GeForce GTX 1080 Ti GPU.

All experiments were performed in PyTorch~\cite{pytorch}. The optimizer of choice was Adam \cite{adam}, with learning rate set to 0.5 empirically. The effect of the adversarial patches on the SS models was evaluated using the mean Intersection-over-Union (mIoU) and mean Accuracy (mAcc) \cite{minaee_image_2020}.


The repository link with the code used for all the experiments was not included for double-blind requirements, but it will be inserted upon request or acceptance.

\vspace{-1.2em}
\paragraph{Datasets}
The experiments in the case of driving images were carried out using the \textit{Cityscapes} dataset \cite{DBLP:conf/cvpr/CordtsORREBFRS16}, a popular benchmark for urban scene SS. The dataset consists of 2975 and 500 high resolution images ($1024\times2048$) for training and validation, respectively. The experiments reported in this paper were conducted on 250 images randomly sampled from the training set. Conversely, the entire validation set was used to evaluate the effectiveness of the patches.

The CARLA simulator was used to provide a 3D virtual scenario. This set of experiments was performed in Town01, one of the built-in maps provided with the simulator, with `CloudyNoon' as the preset weather. To mimic the settings of the Cityscapes dataset, RGB images of size $1024\times2048$, along with their corresponding SS tags, were collected by placing a camera on-board the ego vehicle. 

The SS models trained on Cityscapes had to be fine-tuned to ensure good performance on CARLA images. 600 images for fine-tuning, 100 for validation, and 100 for testing were collected by spawning the ego vehicle at random positions in Town01. Additional details about the fine-tuning can be found in the supplementary material.

To study the effectiveness of the patches in CARLA, the map of Town01 was manually edited to include three billboards. Thus, without-EOT, EOT-based, and the scene-specific attacks were carried out at three different locations within Town01. The datasets for the attacks were collected by spawning the ego vehicle at random locations within the proximity of each billboard to emulate varying viewpoints from which the patch might be captured in the real-world. For each of the three billboards, 150 training images, 100 validation, and 100 test images were gathered. Details about the position and orientation of the billboard and the camera were stored to compute the roto-translations used in the scene-specific attack.

Lastly, to study the effects of adversarial patches in the real world, an additional dataset of 1000 images, hereafter referred to as \textit{Patches-scapes}, was collected by mounting an action camera on the dashboard of a vehicle using a setup similar that of the Cityscapes dataset, and then driving the vehicle within the streets of our city.

\vspace{-1.2em}
\paragraph{Models} 
The attacks studied in this paper were evaluated using DDRNet~\cite{ddrnet_paper}, BiSeNet~\cite{bisenet_paper}, and  ICNet~\cite{icnet_paper}, which represent the state-of-the-art in real-time SS, making them preferable for the use case of autonomous driving. Additionally, PSPNet~\cite{pspnet_paper} was included in the study for the EOT-based attack on the CityScapes dataset, but not for the scene-specific attack on CARLA, since we are interested in real-time performance.



All the models were loaded with the pre-trained weights provided by the authors (further specifications are provided in the supplementary material.
Table \ref{table:models_perf} summarizes the performance of these models on both the Cityscapes and CARLA validation sets. 



\begin{table}
\centering
\resizebox{8.45cm}{!}{%
\begin{tabular}{|c|c|c|}
\hline
model        & \multicolumn{2}{c|}{mIoU / mAcc} \\ \hline
   & cityscapes      & CARLA (val - scene1 - scene2 -scene3)
\\ \hline
ICNet   & 0.78 ~/~0.85       & 0.70~/~0.84 - 0.53~/~0.70 - 0.64~/~0.74 - 0.62~/~0.74 \\ \hline
BiSeNet & 0.69~/~0.78       & 0.47~/~0.69 - 0.47~/~0.69 - 0.61~/~0.74 - 0.47~/~0.73      \\ \hline
DDRNet  & 0.78~/~0.85       & 0.72~/~0.88 - 0.54~/~0.74 - 0.62~/~0.76 - 0.64~/~0.78      \\ \hline
PSPNet  & 0.79~/~0.85       & nd      \\ \hline
\end{tabular}
}

\caption{mIoU and mAcc of the tested models on Cityscapes (pre-trained) and our CARLA dataset (fine-tuned).}
\label{table:models_perf}
\vspace{-1em}
\end{table}

\subsection{EOT-based patches on Cityscapes}
The Cityscapes dataset is used to optimize two types of patches on the same training images, one with EOT and the other without EOT (non-robust). Three different patch sizes are studied: $150\times300$, $200\times400$, and $300\times600$ pixels. 


The non-robust patches (without EOT) were optimized by placing them at the center of the image at each training iteration (i.e., $\eta(\cdot) =$ fixed position) and applying no appearance transformations (i.e., $\Gamma_a =\emptyset$).


Conversely, the robust optimizations with EOT apply multiple digital transformations. $\Gamma_a$ includes only Gaussian noise with standard deviation 5\% of the image range.
$\Gamma$ includes random scaling ($80\% - 120\%$ of the initial patch size) and random translation defined as follows: if $(c_x, c_y)$ is the center of the image, the position of the patch is randomized within the range $(c_x\pm \tilde{r} \cdot \tilde{W}/2~,~c_y\pm \tilde{r} \cdot \tilde{H}/2)$, where $\tilde{r} \in [0,1]$ is a uniform random value.
The translation range was kept limited, rather than considering the full image space, to ensure a greater stability and faster convergence.
The patches were optimized over 200 epochs.

\begin{figure*}[!t]
\centering
\makebox[\columnwidth]{\includegraphics[scale=0.38]{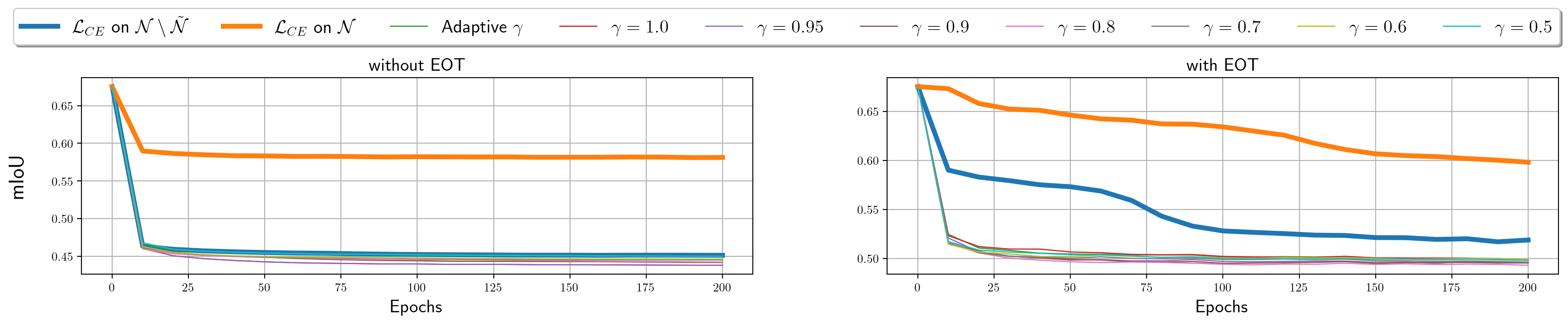}}
\vspace{-1em}
\caption{Comparison of adversarial patch optimizations ($200 \times 400$) on ICNet and Cityscapes using different loss functions: two versions of the standard pixel-wise cross-entropy and our formulation with multiple values of $\gamma$. 
$\mathcal{L}_{CE}$ on $\mathcal{N}$ is the original version used by \cite{nakka_indirect_2019}, while $\mathcal{L}_{CE}$ on $\mathcal{N} \setminus \mathcal{\tilde{N}}$ is an improved version based on the rationale presented in Section \ref{s:method}.}
\label{fig:digital_gamma_ablation}
\end{figure*}

\vspace{-1.2em}
\paragraph{Loss functions analysis}
Figure \ref{fig:digital_gamma_ablation} reports the mIoU obtained by training patches with the pixel-wise CE computed on $\mathcal{N}$ (used by \cite{nakka_indirect_2019}) and $\mathcal{N} \setminus \mathcal{\tilde{N}}$ compared to the extended CE loss proposed in this paper, with $\gamma \in \{ 0.5, 0.6, 0.7, 0.8, 0.95, 1.0, \frac{|\Upsilon|}{|\mathcal{N} \setminus \mathcal{\tilde{N}}|}\}$. Among the models evaluated in the paper, ICNet~\cite{icnet_paper} appears to be most robust on the Cityscapes dataset. Thus, the loss functions are studied by optimizing a $200\times400$ patch with and without EOT on ICNet.


For all the tested values of $\gamma$, our formulation converges to a higher attack effect (i.e., smaller mIoU) with lesser number of epochs than the one based on the pixel-wise CE. Experiments without EOT show that all the compared implementations converge after only $10$ epochs.
In the EOT case, the advantages are even more evident: our proposed formulation converges at almost 25 epochs, while the CE cases still reduce slowly at 200 epochs (nearly $6$ hours of optimization time).

The same study was performed for the scene-specific attack in the CARLA virtual world, and produced similar results, reported in the supplementary material.

\vspace{-1.2em}
\paragraph{Adversarial patch effects.} 
Table \ref{table:results_digital} reports how varying the patch size affects each of the SS models. We used the adaptive $\gamma$ (i.e., $\gamma=\frac{|\Upsilon|}{|\mathcal{N} \setminus \mathcal{\tilde{N}}|}$) that has shown the best overall effect among multiple experimental tests. 

\begin{figure*}[!t]
     \centering
     \begin{subfigure}{0.19\textwidth}
         \centering
         \includegraphics[width=\textwidth]{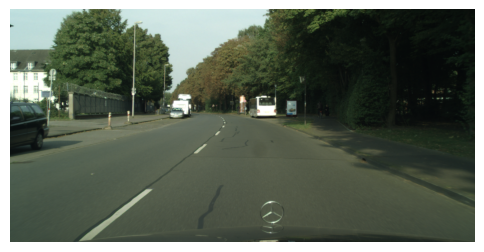}
         \vspace{-1.5em}
         \caption{}
         \label{fig:digital_exps_a}
     \end{subfigure}
     \begin{subfigure}{0.19\textwidth}
         \centering
         \includegraphics[width=\textwidth]{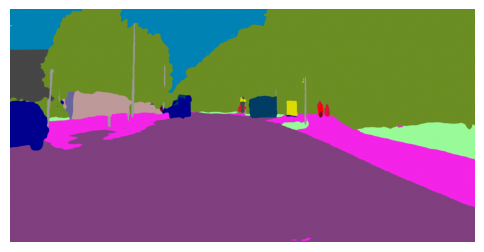}
         \vspace{-1.5em}
         \caption{}
         \label{fig:digital_exps_b}
     \end{subfigure}
     \begin{subfigure}{0.19\textwidth}
         \centering
         \includegraphics[width=\textwidth]{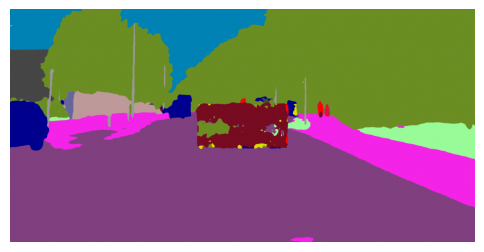}
         \vspace{-1.5em}
         \caption{}
         \label{fig:digital_exps_c}
     \end{subfigure}
     \begin{subfigure}{0.19\textwidth}
         \centering
         \includegraphics[width=\textwidth]{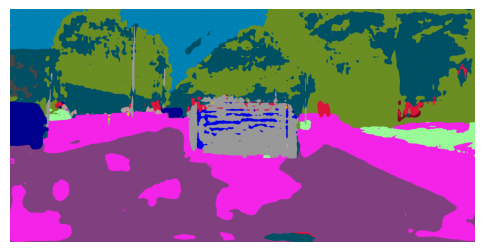}
         \vspace{-1.5em}
         \caption{}
         \label{fig:digital_exps_d}
     \end{subfigure}
     \begin{subfigure}{0.19\textwidth}
         \centering
         \includegraphics[width=\textwidth]{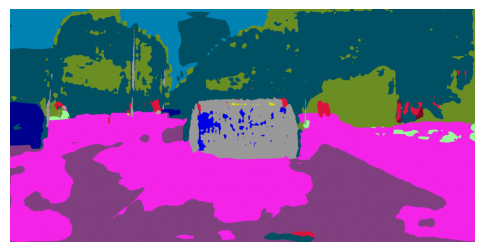}
         \vspace{-1.5em}
         \caption{}
     \end{subfigure}
        \vspace{-1em}
        \caption{Semantic segmentations obtained from BiSeNet with no patch (b), a random patch (c), an EOT-based patch (d), and non-robust patch (without EOT) (e) added into a original image (a) of the Cityscapes validation set.}
        \vspace{-1em}
        \label{fig:digital_exps}
\end{figure*}

Figure \ref{fig:digital_exps} illustrates the effects of the optimized patches on the BiSeNet model.
As expected, the non-robust patches (without EOT) obtain better attack performance with respect to the ones optimized with EOT. This is because the optimization process is simpler when not considering the randomized transformations. However, it is important to note that these patches would not be transferrable to the real world, and are not robust even to simple transformations \cite{2017arXiv171100117G, 2021arXiv210111466N}.


\begin{table*}[!t]
\centering
\small
\begin{tabular}{|c|c|l|c|l|c|l|}
\hline
Model   & \multicolumn{6}{c|}{\begin{tabular}[c]{@{}c@{}}mIoU  |  mAcc \quad(rand / with EOT / without EOT)\end{tabular}}         \\ \hline
        & \multicolumn{2}{c|}{150x300}          & \multicolumn{2}{c|}{200x400}          & \multicolumn{2}{c|}{300x600}          \\ \hline
ICNet   & 0.70 / 0.58 / 0.54 & 0.81 / 0.69 / 0.65 & 0.67 / 0.50 / 0.45 & 0.79 / 0.61 / 0.55 & 0.60 / 0.38 / 0.28 & 0.72 / 0.42 / 0.34 \\ \hline
BiSeNet & 0.63 / 0.45 / 0.39 & 0.74 / 0.61 / 0.55 & 0.61 / 0.29 / 0.21 & 0.71 / 0.43 / 0.34 & 0.54 / 0.19 / 0.05 & 0.65 / 0.31 / 0.15 \\ \hline
DDRNet  & 0.73 / 0.65 /0.55 & 0.82 / 0.76 / 0.64 & 0.71 / 0.59 / 0.42 & 0.80 / 0.69 / 0.50 & 0.65 / 0.45 / 0.09 & 0.73 / 0.53 / 0.19 \\ \hline
PSPNet  & 0.76 / 0.42 / 0.33 & 0.82 / 0.57 / 0.45 & 0.73 / 0.23 / 0.00 & 0.79 / 0.30 / 0.05 & 0.67 / 0.01 / 0.00 & 0.73 / 0.06 / 0.05  \\ \hline
\end{tabular}
\caption{Adversarial patch results on the Cityscapes dataset. Each cell reports the final mIoU obtained with a random patch (no optimization), with EOT, and without EOT.}
\label{table:results_digital}
\end{table*}

\subsection{Scene-specific patches on CARLA}
The scene-specific attack was performed on the same set of models as defined earlier. Each of these models were first fine-tuned on images generated via CARLA. The performance of these fine-tuned models on the CARLA datasets is summarized in Table \ref{table:models_perf}. Please note that the mIoU score is computed as an average of the per-class IoU scores, which, for CARLA, might get to 0 for some classes due to the presence of a few pixels belonging to non-common objects. 

As described in Section \ref{subsection:scene-specific-approach}, the patch is optimized to be adversarial for a specific urban scene by reprojecting it on the attackable 2D surface, which, in this work, is a billboard placed in three different spots in the Town01 map of CARLA. 
This section reports the effect of the scene-specific attack compared against the non-robust (without-EOT) and the EOT-based attacks.
The optimized patch is composed of $150\times300$ pixels, imposing a real-world dimension of $3.75m \times 7.5 m$. Additional experiments on the effect of the real-world dimension of the patch and the number of pixels used are presented in the supplementary material.
For all the following experiments, $\Gamma_a$ includes contrast and brightness changes (both 10\% of the image range), and Gaussian noise (standard deviation 10\% of the image range). 

\begin{table*}[!t]
\centering
\resizebox{\textwidth}{!}{
\begin{tabular}{|c|c|l|c|l|c|l|}
\hline
Model   & \multicolumn{6}{c|}{\begin{tabular}[c]{@{}c@{}}mIoU  |  mAcc \quad(rand / without EOT / EOT / scene-specific)\end{tabular}}         \\ \hline
        & \multicolumn{2}{c|}{Scene1}          & \multicolumn{2}{c|}{Scene2}          & \multicolumn{2}{c|}{Scene3}          \\ \hline
ICNet   & 0.51 / 0.51 / 0.49 / 0.48 & 0.60 / 0.60 /0.56 / 0.54 & 0.64 / 0.64 / 0.61 / 0.61 & 0.74 / 0.74 / 0.73 / 0.73 & 0.63 / 0.63 / 0.59 / 0.59 & 0.76 / 0.76 / 0.73 / 0.74  \\ \hline
BiSeNet & 0.44 / 0.42 / 0.36 / 0.31 & 0.63 / 0.61 / 0.55 / 0.49 & 0.60 / 0.60 / 0.58 / 0.58 & 0.76 / 0.75 / 0.74 / 0.74 & 0.47 / 0.46 / 0.46 / 0.45 & 0.74 / 0.73 / 0.73 / 0.73 \\ \hline
DDRNet  & 0.51 / 0.50 / 0.46 / 0.46 & 0.70 / 0.69 / 0.69 / 0.69 & 0.62 / 0.62 / 0.52 / 0.49 & 0.76 / 0.75 / 0.71 / 0.66 & 0.65 / 0.65 / 0.58 / 0.59 & 0.78 / 0.78 / 0.76 / 0.76 \\ \hline
\end{tabular}
}
\vspace{-0.5em}
\caption{Adversarial patch results on the three scene CARLA datasets. The Table reports the mIoU and mAcc obtained with random, non-robust (without EOT), EOT-based and scene-specific patches.}
\vspace{-0.5em}
\label{table:results_carla}
\end{table*}

\begin{figure*}[!t]
     \centering
     \begin{subfigure}{0.19\textwidth}
         \centering
         \includegraphics[width=\textwidth]{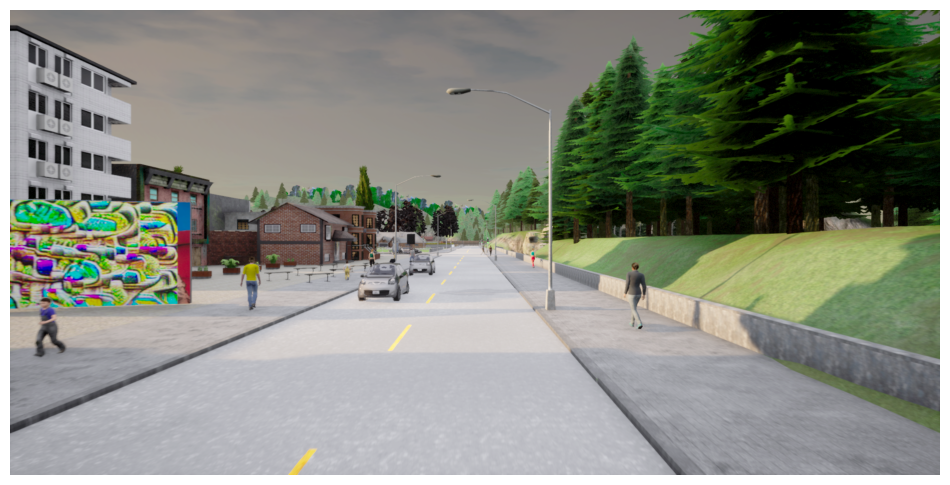}
         \vspace{-1.5em}
         \caption{}
         \label{fig:digital_exps_a}
     \end{subfigure}
     \begin{subfigure}{0.19\textwidth}
         \centering
         \includegraphics[width=\textwidth]{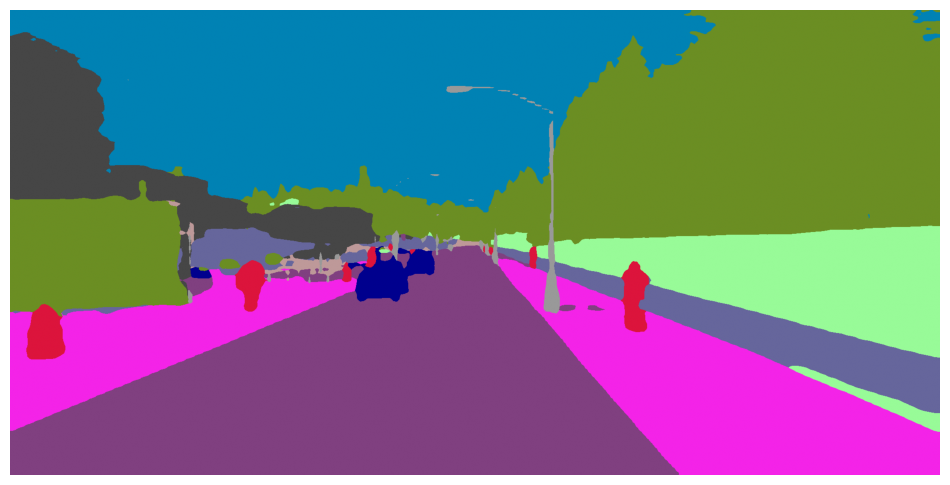}
         \vspace{-1.5em}
         \caption{}
         \label{fig:digital_exps_b}
     \end{subfigure}
     \begin{subfigure}{0.19\textwidth}
         \centering
         \includegraphics[width=\textwidth]{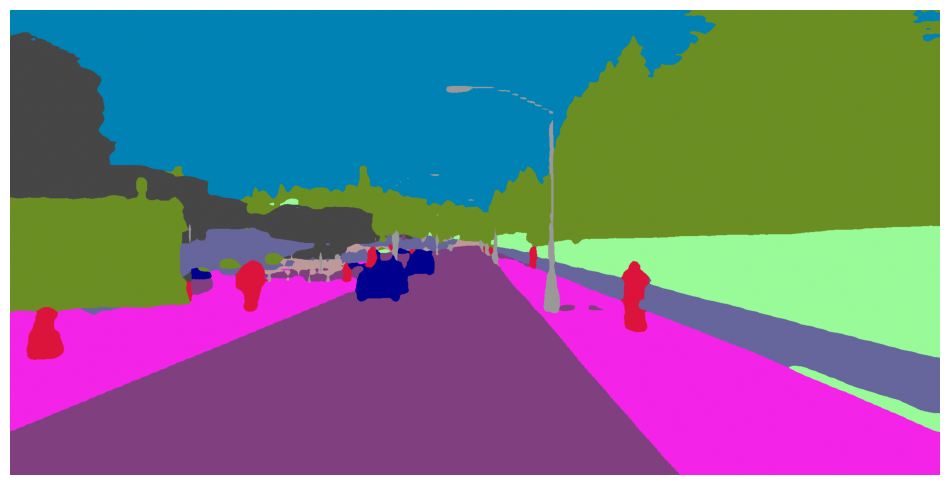}
         \vspace{-1.5em}
         \caption{}
         \label{fig:digital_exps_c}
     \end{subfigure}
     \begin{subfigure}{0.19\textwidth}
         \centering
         \includegraphics[width=\textwidth]{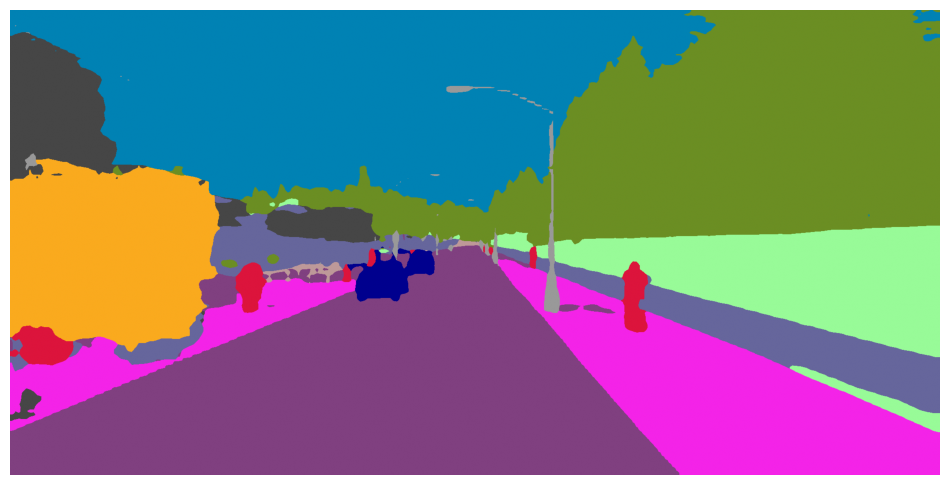}
         \vspace{-1.5em}
         \caption{}
         \label{fig:digital_exps_d}
     \end{subfigure}
     \begin{subfigure}{0.19\textwidth}
         \centering
         \includegraphics[width=\textwidth]{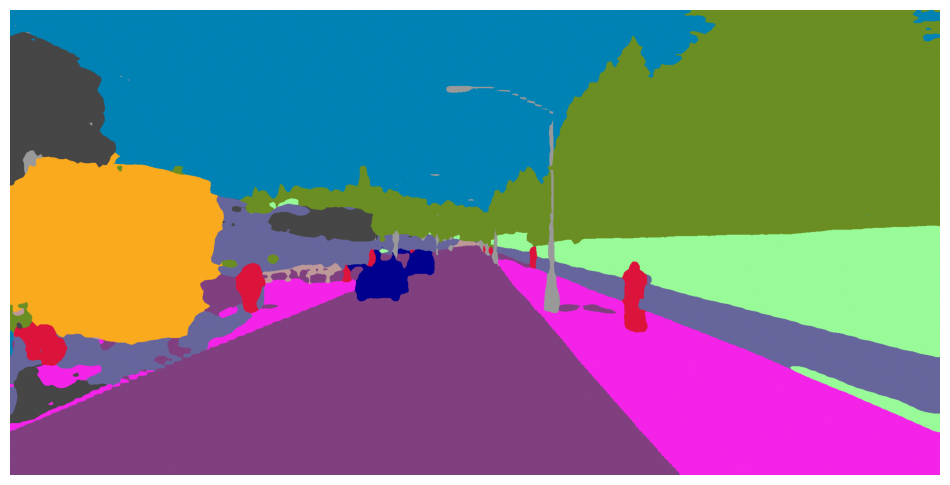}
         \vspace{-1.5em}
         \caption{}
     \end{subfigure}
        \vspace{-1em}
        \caption{(a) is an image extracted from the scene-1 test dataset augmented with a scene-specific patch optimized on DDRNet, while (e) is its corresponding SS; (b), (c), and (d) are predictions obtained by augmenting the same test image with a random, non-robust and EOT-based patches, respectively.}
        \label{fig:carla_exps}
        \vspace{-1em}
\end{figure*}

Since the objective of this work is to craft RWAEs, the performance of the attacks is evaluated by measuring the mIoU and mAcc scores on additional scene-specific datasets. These additional datasets are produced by collecting the same images of the validation set of each scene-specific dataset, but with a single major modification: the billboard object is modified in the \textit{Unreal Editor} \cite{unrealengine} by applying the optimized patch as a \emph{decal} object, which is a way to stick an image on a surface in the virtual environment. This method should provide a simulated real-world application of the patches, since they are no more applied directly on the image, but the image itself includes the patched billboard. 

Table \ref{table:results_carla} summarizes the results obtained on these three additional scene-specific datasets, with a random patch, a non-robust patch, an EOT-based patch, and a scene-specific patch. Figure \ref{fig:carla_exps} shows a comparison of all the discussed attacks on DDRNet.

For almost all the combinations of scene and network, the scene-specific attack outperforms the EOT-based attack, confirming that the scene-specific attack, for this kind of problems, is a better alternative to the EOT formulation for the placement of the patch within the image. The only case where the the two attacks show comparable performance is for scene 3, where the billboard is almost perpendicular to the camera plane, allowing the EOT method to cover realistic patch placing functions. 

It is also worth noting that SS models are rather robust to adversarial patch attacks in general. Although it is possible to craft adversarial patches that cause a section of the image to be wrongly segmented, it tends to be more difficult than fooling models for tasks such as classification. 
Additional details are reported in the supplementary material.

\subsection{Real-world patches}
\begin{figure}[!t]
     \centering
     \begin{subfigure}{0.23\textwidth}
         \centering
         \includegraphics[width=\textwidth]{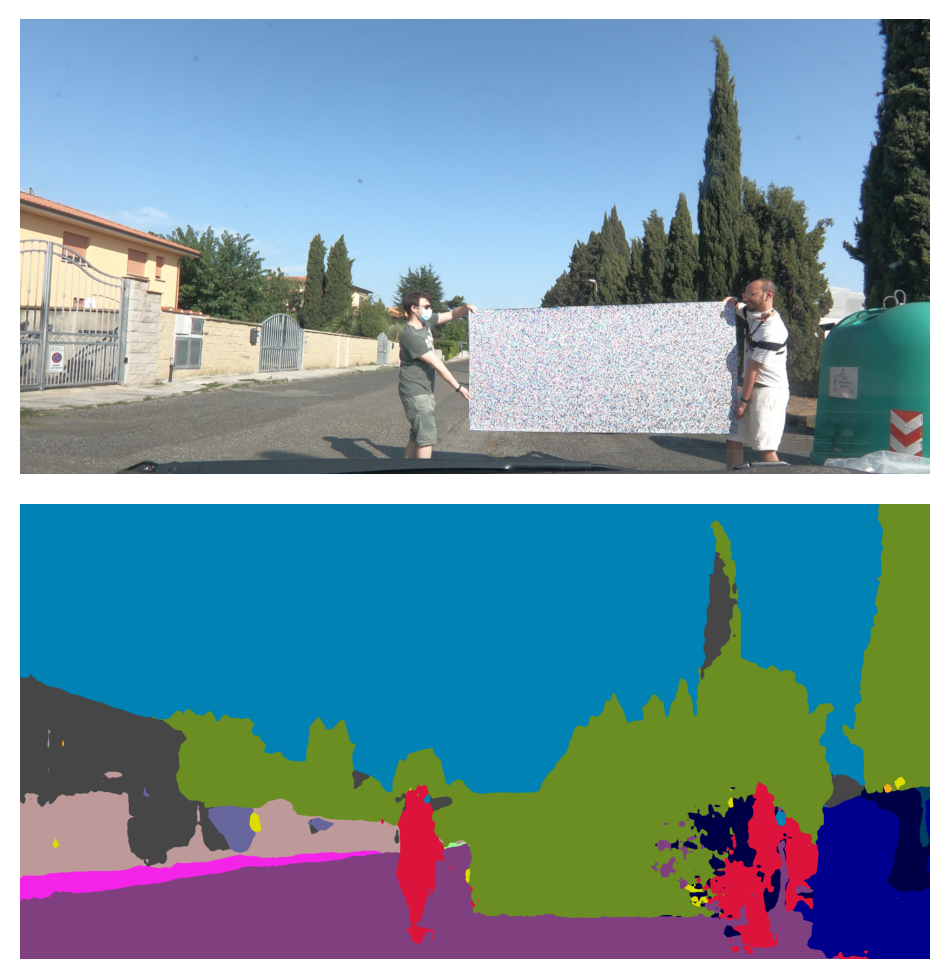}
         \vspace{-1.5em}
         \label{fig:realworld_a}
     \end{subfigure} 
     \begin{subfigure}{0.23\textwidth}
         \centering
         \includegraphics[width=\textwidth]{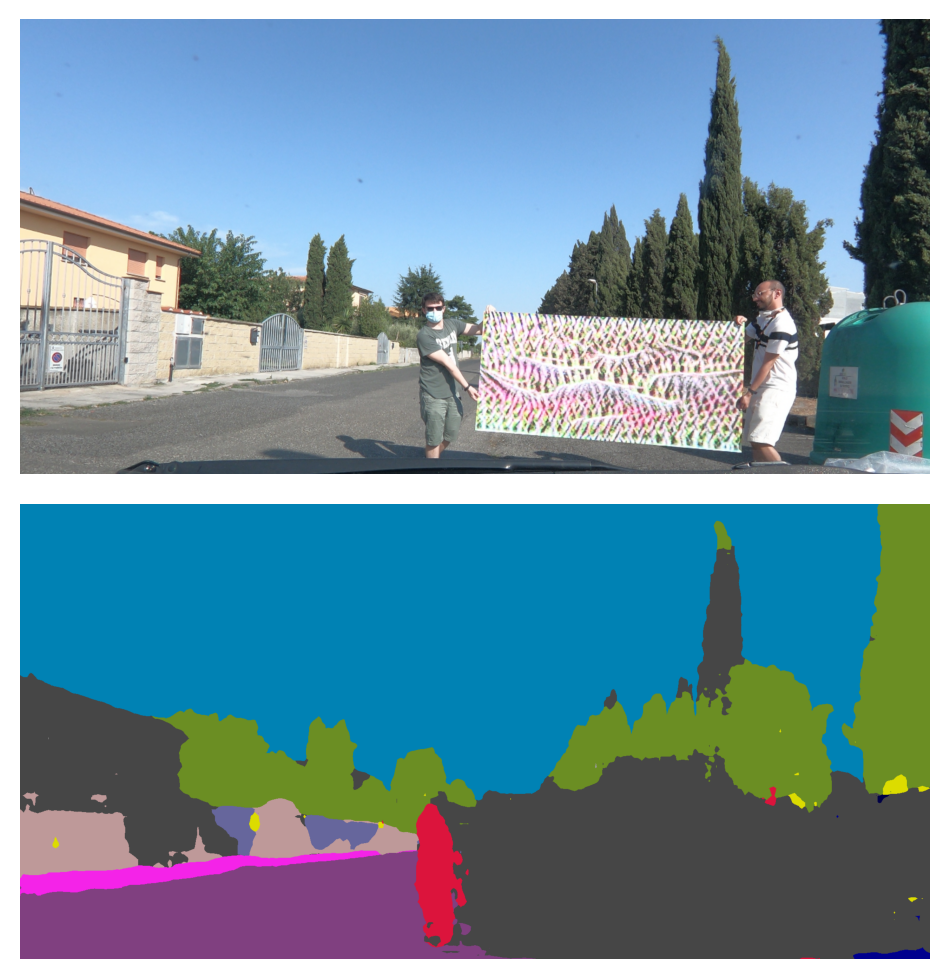}
         \vspace{-1.5em}
         \label{fig:realworld_b}
     \end{subfigure}
        \caption{Real-world predictions on ICNet obtained with a printed random patch (left) and an optimized patch (right).}
        \label{fig:realworld}
        \vspace{-1em}
\end{figure}

In order to prove that the proposed pipeline can be used for a real-world attack, we use the \textit{Patches-scapes} dataset (described in Section \ref{s:exp_setup}) to craft an adversarial real-world patch using the EOT-based patch attack. This section presents the results of an attack in Figure \ref{fig:realworld}. 
Although the presence of the optimized patch does alter a significant area of the predicted SS (while the random patch does not), portions of the image far from its position are not affected. Furthermore, the attack performance decreases as we move the patch away from the camera (details provided in the supplementary material.
The patch is optimized for 200 epochs on the pre-trained version of ICnet (since it showed good performance on the Patches-scapes dataset) and printed as a $1m \times 2m$ poster.

Testing adversarial patches for autonomous driving in the real world poses a series of difficulties that heavily limited the tests. First, it is not easy to find a urban corner with good prediction accuracy, and which is not crowded with moving vehicles (which might be dangerous). Second, the patch must be printed in the highest resolution possible on a large rigid surface, which might get expensive. Furthermore, since weather conditions are not controllable and change throughout the day, results can diverge from what is expected. 

The scene-specific attack, which requires additional geometric information, could not be implemented at the time of writing, but will be considered in future work.

\section{Conclusions and future work} \label{s:conclusions}

This paper presented an extensive study of the adversarial robustness of semantic segmentation models. This was accomplished by extensively evaluating the effect introduced by adversarial patches, to investigate the limits of real-world attacks for segmentation neural networks in an autonomous driving scenario.
Carrying out the investigations with increasingly ``real-world" benchmarks, we studied the effect of non-robust and EOT-based patches on the Cityscapes dataset, on a virtual 3D scenario, and in a real-world setting. We also introduced a new method called \textit{scene-specific attack}, which improves the EOT formulation for a more realistic and effective patch placement. 

The novel loss function introduced in the paper enabled to advance the state-of-the-art for adversarial patches optimization methods, as it proved to be a more general and efficient alternative to the classic cross-entropy function for this kind of problems.

This exhaustive set of experiments practically opens to a new point of view for studying SS models in autonomous driving. Although the proposed attacks were able to reduce the baseline model accuracy, the SS models proved to be somehow robust to real-world patch-based attacks. This was especially noticeable when the tests were performed in more realistic settings using CARLA and the real world, where, in most cases, the patch only affected the proximity of the attacked surface.

Nevertheless, this is a promising result, since it shows how the prediction provided by these models is not easily corruptible, especially in real-world scenarios. This is in contrast with previous work on patch-based adversarial attacks against classification and object detection models.

Future work will further investigate the robustness properties of these models, introducing defense mechanisms and trying to enhance the robustness of SS model by adding a temporal dimension. 
\vfill

{\small
\bibliographystyle{ieee_fullname}
\bibliography{realworld}
}

\end{document}